\documentclass[lettersize,journal]{IEEEtran}

\usepackage{amsmath,amssymb,amsfonts}
\usepackage{algorithm}
\usepackage{algorithmic}
\usepackage{array}
\usepackage{booktabs}
\usepackage{multirow}
\usepackage{graphicx}
\usepackage{subfig}
\usepackage{textcomp}
\usepackage{xcolor}
\usepackage{cite}
\usepackage{pifont}
\usepackage[T1]{fontenc}
\usepackage[utf8]{inputenc}
\usepackage{graphicx}
\graphicspath{{fig/}}
\usepackage{mathtools}
\usepackage{bm} % bold math
\usepackage[hidelinks]{hyperref}
\newcommand{\pmfmt}[2]{\ensuremath{#1^{\pm #2}}}

\newcommand{\ulpm}[2]{\ensuremath{\underline{#1}^{\pm #2}}}

\newcommand{\bpm}[2]{\ensuremath{\mathbf{#1}^{\boldsymbol{\pm #2}}}}
\newcommand{\bval}[1]{\ensuremath{\mathbf{#1}}}

\begin{document}

\title{Pose-Guided Residual Refinement for Interpretable Text-to-Motion Generation and Editing}

\author{Sukhyun Jeong and Yong-Hoon Choi,~\IEEEmembership{Member,~IEEE}%
\thanks{Date of submission: December 13, 2025. This work was supported by the Korea Agency for Infrastructure Technology Advancement (KAIA) grant funded by the Ministry of Land, Infrastructure and Transport under the Smart Building R\&D Program (Grant No. RS-2025-02532980); by the National Research Foundation of Korea (NRF) grant funded by the Korea government (MSIT) (RS-2025-16069933); and by the Research Grant of Kwangwoon University in 2024.}%
\thanks{(Corresponding author: Yong-Hoon Choi.)}%
\thanks{The authors are with the Fintech and AI Robotics (FAIR) Laboratory, the School of Robotics, Kwangwoon University, Nowon-gu, Seoul 01897, South Korea (e-mail: jayze3736@gmail.com; yhchoi@kw.ac.kr).}%
}

\IEEEtitleabstractindextext{%
\begin{abstract}
Text-based 3D motion generation aims to automatically synthesize diverse motions from natural-language descriptions to extend user creativity, whereas motion editing modifies an existing motion sequence in response to text while preserving its overall structure. Pose-code-based frameworks such as CoMo map quantifiable pose attributes into discrete pose codes that support interpretable motion control, but their frame-wise representation struggles to capture subtle temporal dynamics and high-frequency details, often degrading reconstruction fidelity and local controllability. To address this limitation, we introduce pose-guided residual refinement for motion (PGR²M), a hybrid representation that augments interpretable pose codes with residual codes learned via residual vector quantization (RVQ). A pose-guided RVQ tokenizer decomposes motion into pose latents that encode coarse global structure and residual latents that model fine-grained temporal variations. Residual dropout further discourages over-reliance on residuals, preserving the semantic alignment and editability of the pose codes. On top of this tokenizer, a base Transformer autoregressively predicts pose codes from text, and a refine Transformer predicts residual codes conditioned on text, pose codes, and quantization stage. Experiments on HumanML3D and KIT-ML show that PGR²M improves Fréchet inception distance and reconstruction metrics for both generation and editing compared with CoMo and recent diffusion- and tokenization-based baselines, while user studies confirm that it enables intuitive, structure-preserving motion edits. Implementation code, demo, and pretrained models are publicly available at \url{https://github.com/jayze3736/PGR2M}.
\end{abstract}

\begin{IEEEkeywords}
Residual vector quantization (RVQ), motion editing, text-to-motion, pose code, transformer.
\end{IEEEkeywords}}

\maketitle
\IEEEdisplaynontitleabstractindextext
\IEEEpeerreviewmaketitle

%\markboth{IEEE Transactions on Human-Machine Systems,~Vol.~XX, No.~X, Month~2025}%

\section{Introduction}\label{sec:intro}

\IEEEPARstart{T}{ext}-to-motion (T2M) generation aims to predict 3D human joint
trajectories from natural-language descriptions and can serve as a key
enabling technology in virtual reality (VR), animation, and robotics by
reducing manual authoring costs and increasing productivity. Early
studies represented an entire motion sequence as a continuous latent
sequence composed of short motion segments and trained models to predict
this latent representation from text \cite{ref1}.

Motivated by the analogy between language and motion, subsequent work
began to treat human motion as a sequence of discrete tokens, analogous
to words in a sentence. To this end, continuous motion sequences are
first mapped into a vector-quantized latent space using a VQ-VAE model
\cite{ref2}, and short motion clips are represented as tokens from a learned
motion codebook \cite{ref3,ref4,ref5}. This discretization enables the direct
application of Transformer-based language models \cite{ref6} to motion
modeling, yielding simple yet effective T2M generators. In parallel,
diffusion-based models generate motion by injecting Gaussian noise into
either the joint space or a learned latent space and gradually denoising
it under text conditioning, thereby synthesizing fine motion details
through multi-step refinement \cite{ref7,ref8,ref9}.

Beyond pure generation, recent research has proposed frameworks that
support editing of an existing motion sequence in response to a user's
textual instruction \cite{ref7}, \cite{ref10}, \cite{ref11}. In these approaches,
the editing prompt and the original motion are jointly used as
conditions for the generative model, which then resamples a modified
motion that reflects the requested local changes (e.g., modifying
specific body parts, adjusting speed or rhythm). Unlike generation tasks
that primarily aim to produce diverse samples, motion editing requires
precise, localized control while preserving the overall structure of the
original motion.

CoMo \cite{ref11} points out that editing by fully regenerating motion
through a generative model often fails to maintain consistency between
the original and edited motions. To mitigate this issue, CoMo introduces
interpretable pose codes that compactly encode high-level pose
states---such as joint angles, inter-joint distances, and relative
orientations---at each frame. A frame-wise pose is represented as a
combination of pose codes selected from multiple pose categories,
enabling users to perform local edits on specific frames or joints while
preserving the global structure of the original motion. In this sense,
pose codes serve as key units that simultaneously ensure structural
consistency and intuitive editability.

However, pose codes are essentially static, frame-wise descriptors and
thus exhibit structural limitations in capturing subtle temporal
variations across consecutive frames and high-frequency motion
characteristics. While they effectively compress the spatial
configuration of a single pose, they are less suited to representing
fine-grained motion dynamics, such as accumulated micro-movements or
precise variations in speed and rhythm over time. As a result, motion
reconstructed or edited using only pose codes tends to preserve global
pose structure and coarse motion patterns, but often lacks detailed,
high-frequency motion nuances.

To address this limitation, we build on the interpretability and
editability of CoMo's pose codes and introduce pose-guided residual
refinement for motion (PGR²M), which augments pose codes with residual
codes learned via residual vector quantization (RVQ). In the proposed
framework, pose codes are responsible for the coarse, global motion
structure, whereas residual codes model fine temporal variations and
high-frequency motion details that are not captured by pose codes alone.
By combining pose codes with RVQ-based residual codes, PGR²M enables
interpretable and controllable motion editing while significantly
improving the fidelity of reconstructed and generated motions.

The remainder of this paper is organized as follows. Section II reviews
related work on vector-quantized motion representations and
text-to-motion generation and editing. Section III introduces the
proposed pose-guided residual refinement framework, including the
pose-guided RVQ tokenizer and the base and refine Transformers for
text-conditioned motion generation and editing. Section IV details the
training objectives for the tokenizer and the Transformers. Section V
presents experimental results on the HumanML3D and KIT-ML datasets
together with ablation studies and user evaluations. Section VI
concludes the paper and discusses directions for future research.

\section{Related Work}\label{sec:related}

The vector-quantized variational autoencoder (VQ-VAE) \cite{ref2} is a
discrete representation model that has been widely applied in various
domains, including image and speech synthesis \cite{ref12}, \cite{ref13}. It
maps continuous input data into latent vectors through an encoder and
then quantizes each latent vector to the nearest code in a learnable
codebook composed of discrete codes. The codebook is trained to
approximate the data distribution, and each code functions as a token
that represents a characteristic pattern or structure. Consequently,
VQ-VAE enables the modeling of data distributions in a discrete code
space instead of a continuous latent space.

T2M-GPT \cite{ref4} employs VQ-VAE as a motion tokenizer: continuous motion
sequences are converted into sequences of discrete tokens drawn from a
motion codebook, and a Transformer-based model is trained to
autoregressively predict the motion token sequence conditioned on a
given text description.

MotionGPT \cite{ref5} focuses on the correspondence between motion and
natural language from a linguistic perspective and proposes a
language-model-based framework that treats motion tokens and text tokens
as a unified sequence. In this framework, not only text-to-motion
generation but also motion-to-text generation, which produces
natural-language descriptions from motion sequences, are formulated as
conditional sequence prediction problems of a language model. The model
is trained on bidirectional tasks in which both motion tokens and text
tokens appear jointly in the input and output sequences, thereby
enabling various motion-related tasks to be handled in a unified manner
within a single model.

MDM \cite{ref7} is a diffusion-based motion generation model that directly
operates in the joint space of motion sequences and, in particular,
supports conditional editing of upper-body motions. During the editing
process, the model masks joint features corresponding to the lower body
so that they are excluded from the diffusion updates, and re-estimates
only the joint features related to the upper body conditioned on an
editing prompt, thereby modifying the motion.

FineMoGen \cite{ref10} refines the spatiotemporal constraints encoded in
text conditions by decoupling motion modeling into separate modules for
temporal and spatial features and generating motion while precisely
incorporating spatiotemporal features derived from the text. For motion
editing, the user's editing prompt is fed into a large language model
(LLM) to modify a fine-grained caption of the original motion; the
updated caption is then used as a new text condition to regenerate the
motion and obtain the edited result.

GraphMotion \cite{ref9} is a diffusion-based motion generation model that
introduces a graph-based text embedding to better capture textual
conditions. It constructs a graph over words in a sentence and employs a
graph neural network (GNN) to model their semantic relationships and
dependencies, producing a structured text representation. This
representation is used as a condition in the diffusion process, enabling
the model to generate motions that more accurately reflect the action
semantics specified in the text.

CoMo \cite{ref11}, inspired by PoseScript \cite{ref14}, represents a pose as a
combination of multiple pose codes. Each pose code denotes a high-level
state that can be quantitatively derived from a pose, such as the amount
of bending of the right arm, distances between specific joints, or
relative orientations. The pose at a single frame is thus described as a
combination of pose codes selected from multiple pose categories. This
representation makes motions expressed in pose codes intuitively
interpretable, and users can directly edit the pose in a desired manner
by modifying pose codes over specific temporal segments. Such a discrete
pose representation enables intuitive motion editing while preserving
consistency between the original and edited motions and further provides
an interpretable representation space that can be integrated with
language-based models.

However, because pose codes are fundamentally static, frame-wise
descriptors, they have inherent limitations in capturing subtle pose
changes across frames as well as high-frequency motion components such
as variations in speed and rhythm. Therefore, in this work, we aim to
preserve the interpretability and editability of CoMo while introducing
residual vector quantization (RVQ) to model fine-grained feature
information that is not captured by pose codes in the form of residual
codes. By assigning the basic motion structure to pose codes and
representing subtle inter-frame pose changes and temporal variations
with residual codes, our goal is to enable more detailed and natural
motion representations.

\begin{figure*}[!t]
\centering
% \includegraphics[width=\textwidth]{figures/fig1}
% \fbox{\parbox[c][2.2in][c]{\textwidth}{\centering \textit{(Insert Fig.~1 here)}}}
%\includesvg[width=\textwidth]{fig/fig_1_PG_tokenizer}
\includegraphics[width=\textwidth]{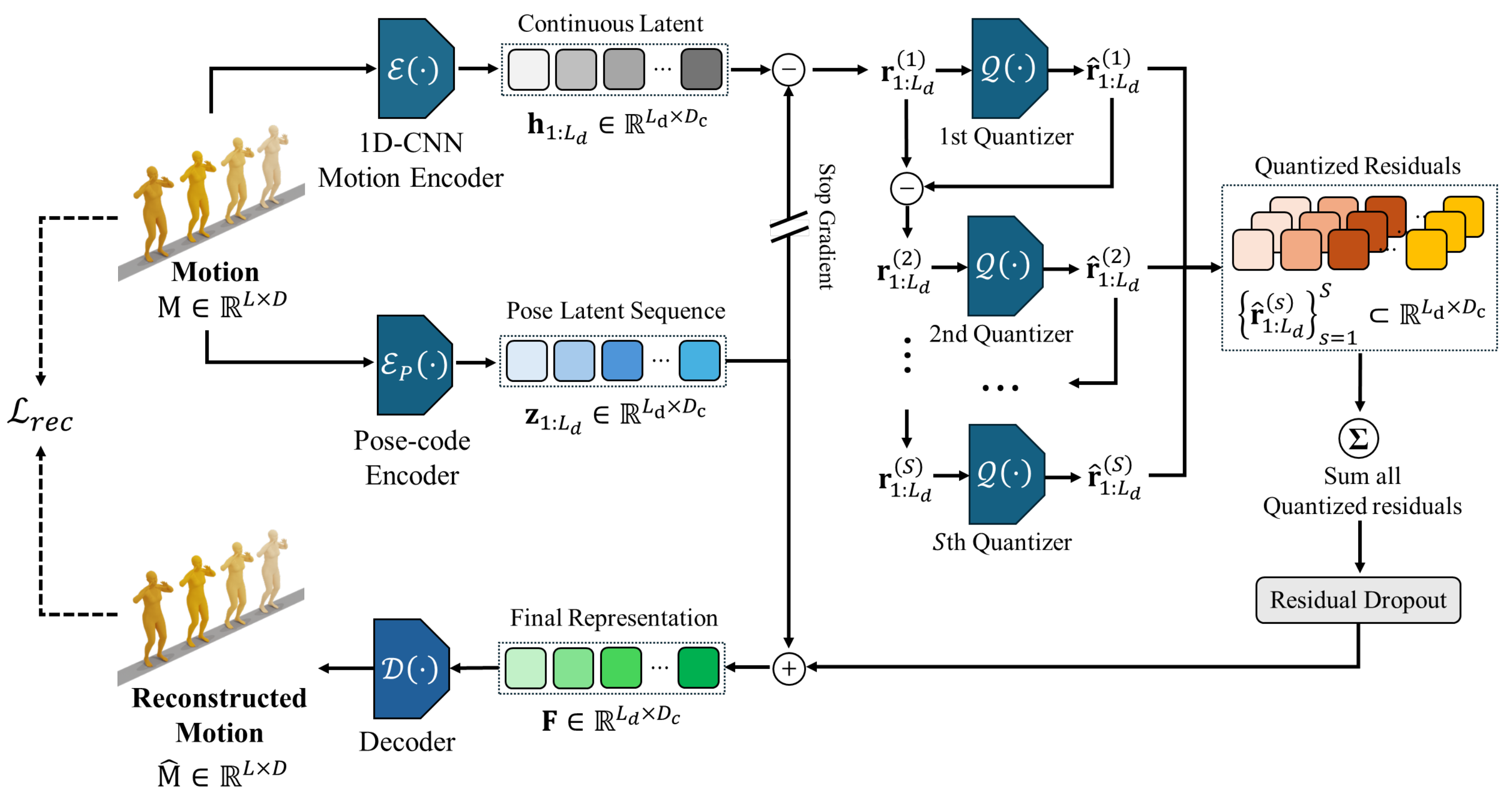}
\caption{Architecture of the proposed pose-guided RVQ tokenizer. The pose code encoder produces interpretable pose latents. The 1D CNN motion encoder produces continuous motion latents. Their difference is progressively quantized by RVQ to obtain residual codes, and residual dropout stochastically masks residuals during training.}
\label{fig:overall}
\end{figure*}

\section{Method}\label{sec:method}

The proposed framework consists of a representation learning backbone
and two text-conditioned predictors. At its core, we learn a joint
representation of motion using pose codes and residual codes through a
pose-guided RVQ tokenizer. On top of this tokenizer, a base Transformer
predicts pose codes from text, and a refine Transformer further predicts
residual codes conditioned on both text and pose codes, yielding the
final motion representation.

\subsection{Pose-Guided RVQ Tokenizer}\label{a.-pose-guided-rvq-tokenizer}

We propose a pose-guided RVQ tokenizer that models the motion space by
decomposing it into two complementary components: an interpretable
pose-code representation that captures the overall motion structure, and
residual codes that refine fine-grained motion details. The overall
architecture is illustrated in Fig. 1.

Let \(\mathbf{M} \in \mathbb{R}^{L \times D}\) denote an input motion
sequence of length \(L\) with \(D\)-dimensional joint features per
frame. The sequence is routed to two parallel encoders. Because pose
codes encode only downsampled, per-frame static poses, they have limited
access to the temporal context of neighboring frames. To mitigate this
limitation, we incorporate a 1D CNN--based motion encoder \cite{ref4} that
extracts features from continuous pose sequences while taking local
temporal dynamics into account.

The pose-code encoder \(\mathcal{E}_{p}\) converts poses in the motion
sequence into pose codes and produces a sequence of pose latent
representations. In parallel, the 1D CNN motion encoder \(\mathcal{E}\)
outputs a continuous latent representation of the motion that reflects
local temporal variation. The difference between the pose latents and
the continuous latents is modeled as a residual and is progressively
quantized by multiple RVQ stages to obtain residual codes. These
residual codes are added back to the pose latent representation to form
the final fused latent representation \(\mathbf{F}\). The decoder then reconstructs the
motion sequence as \(\widehat{\mathbf{M}} = \mathcal{D}(\mathbf{F}) \in \mathbb{R}^{L \times D}\).

\begin{figure*}[!t]
\centering
\begin{minipage}[t]{0.49\textwidth}
  \centering
  \includegraphics[width=\linewidth]{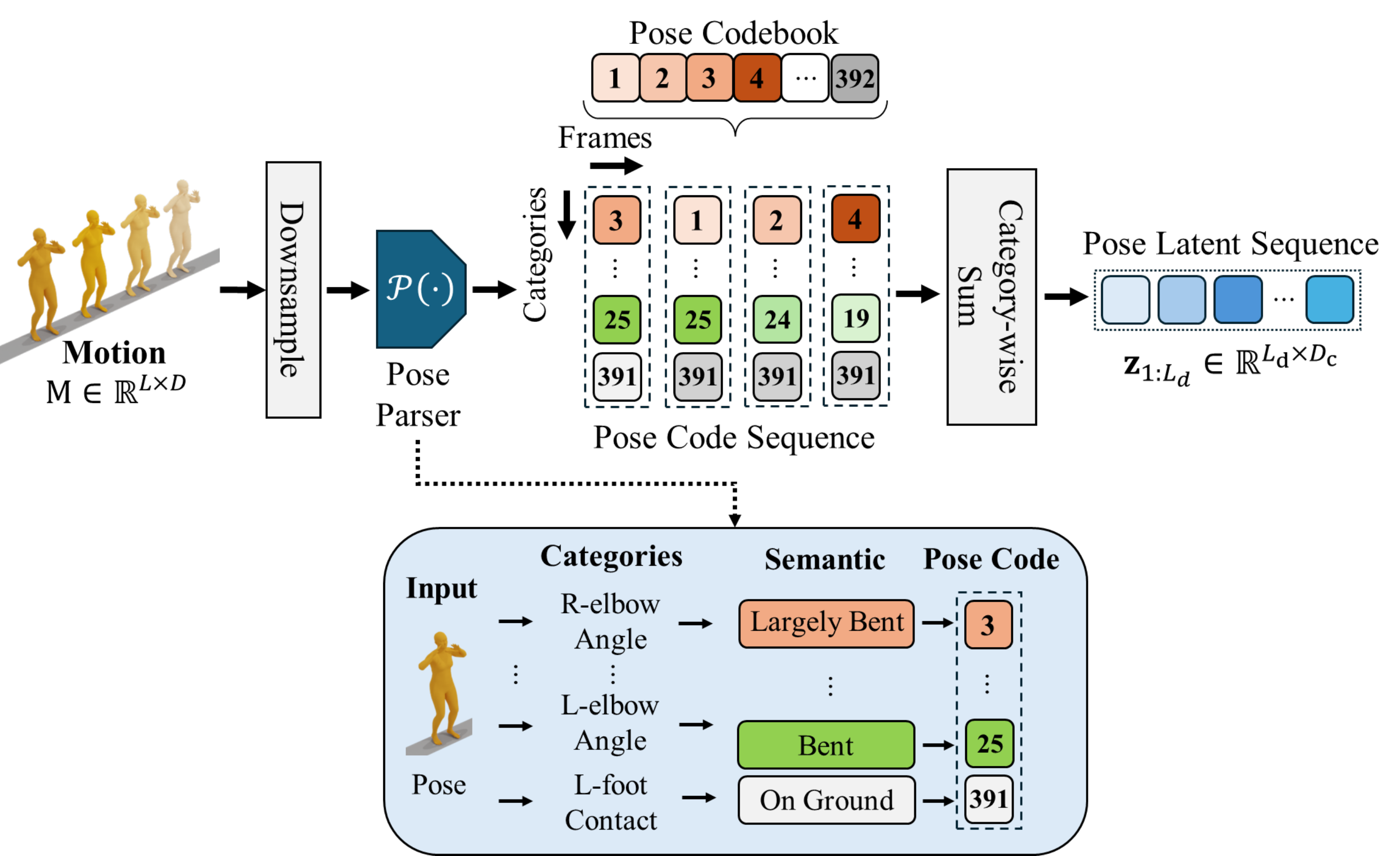}
  \caption{Architecture of the pose code encoder. Each frame is mapped to a multi-hot vector over pose categories by the pose parser, and the pose latent is formed by a linear combination of activated pose codes.}
  \label{fig:pose_encoder}
\end{minipage}\hfill
\begin{minipage}[t]{0.49\textwidth}
  \centering
  \includegraphics[width=\linewidth]{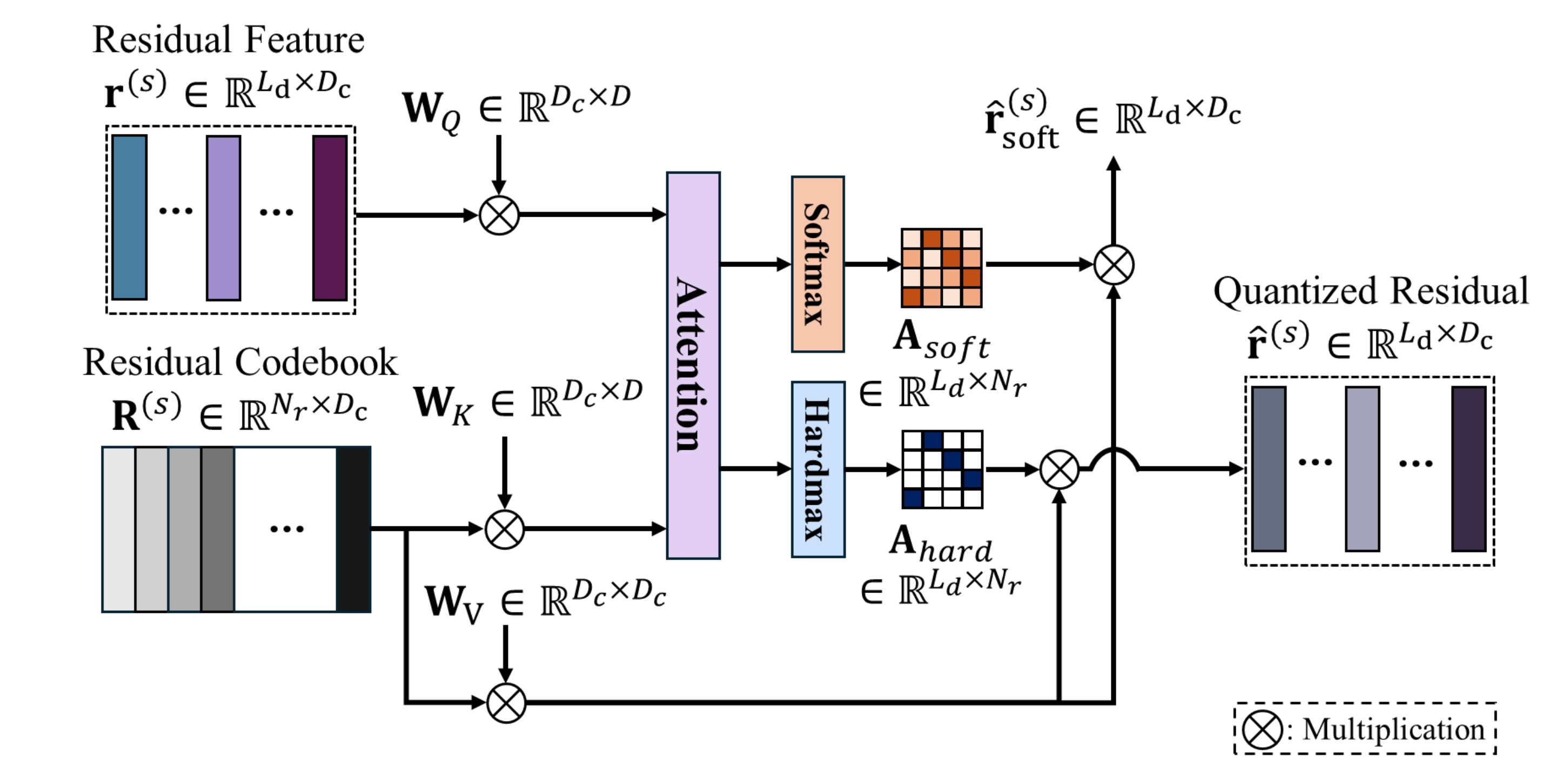}
  \caption{Attention-based quantization process. Queries from the input residual and keys from the codebook entries yield soft and hard selections, producing quantized residuals that are used for reconstruction and for stable training.}
  \label{fig:attn_vq}
\end{minipage}
\end{figure*}

% \newpage

\subsection{Pose Code Encoder}\label{b.-pose-code-encoder}

The pose code encoder, originally introduced as the motion encoder in
\cite{ref11}, consists of a pose parser \(\mathcal{P}\) \cite{ref14} and a pose
codebook
\(\mathbf{C} = \left\{ \mathbf{c}_{n} \right\}_{n = 1}^{N} \subset \mathbb{R}^{D_{c}}\),
as illustrated in Fig. 2.  Here, \(D_{c}\) denotes the dimensionality of
each pose code. Following CoMo \cite{ref11}, we adopt \(N = 70\) pose
categories that describe high-level pose attributes such as joint
angles, pairwise joint distances, relative positions and orientations,
and contact states with the ground.

Given a motion sequence \(\mathbf{M}\) of length \(L\), we first
downsample it with stride \(l\) to obtain a sequence of poses
\(\mathbf{M}_{d} = \left\{ \mathbf{p}_{i \times l} \right\}_{i = 1}^{L_{d}} \subset \mathbb{R}^{D}\),
where \(D\) is the dimensionality of a pose vector and \(L_{d} = L/l\)
is the length of downsampled sequence. For each pose
\(\mathbf{p}_{i \times l}\), the pose parser \(\mathcal{P}\) determines
which pose codes in the codebook $\mathbf{C}$ are activated and outputs an
\(N\)-dimensional binary indicator vector whose \(n\)-th element is
given by

\begin{equation}
{\mathbf{z}_{1:L_{d}} = \left\{ \left\{ \mathcal{P}\left( \mathbf{c}_{n},\ \mathbf{p}_{i \times l} \right) \right\}_{n = 1}^{N} \right\}}_{i = 1}^{L_{d}},
\tag{1}\label{eq:1}
\end{equation}

\noindent where
\(\mathcal{P}\left( \mathbf{c}_{n},\ \mathbf{p}_{i \times l} \right) \in \{ 0,1\}\)
indicates whether pose code \(\mathbf{c}_{n}\) is activated for pose
\(\mathbf{p}_{i \times l}\). Thus, each frame is represented as a
multi-hot vector over the \(N\) pose codes. By arranging the pose
codebook as a matrix \(\mathbf{C} \in \mathbb{R}^{D_{c} \times N}\), the
sequence of pose latent representation is obtained as a linear
combination of pose codes weighted by these binary indicators:

\begin{equation}
\begin{aligned}
\widehat{\mathbf{z}}_{1:L_{d}}
&= \left\{\mathbf{C}\,\mathbf{z}_i\right\}_{i=1}^{L_{d}} \\
&= \left\{\sum_{n=1}^{N} \mathcal{P}\!\left(\mathbf{c}_n, \mathbf{p}_{i\times l}\right)\,\mathbf{c}_n\right\}_{i=1}^{L_{d}}
\in \mathbb{R}^{L_{d} \times D_{c}}.
\end{aligned}
\tag{2}\label{eq:2}
\end{equation}

The sequence \({\widehat{\mathbf{z}}}_{1:L_{d}}\) serves as the pose
latent representation used by the subsequent modules.

\subsection{Motion Encoder and Motion Decoder}\label{c.-motion-encoder-and-motion-decoder}

We adopt the motion encoder and decoder architecture proposed in T2M-GPT
\cite{ref4}. The encoder \(\mathcal{E}\) takes an input motion sequence
\textbf{M} and extracts local motion features, producing a continuous
latent sequence
\(\mathcal{E}\left( \mathbf{M} \right) = \left\{ \mathbf{h}_{i} \right\}_{i = 1}^{L_{d}},\ \ \mathbf{h}_{i} \in \mathbb{R}^{D_{c}}\),
which can be written as
\(\mathcal{E}\left( \mathbf{M} \right) \in \mathbb{R}^{L_{d} \times D_{c}}\).
The decoder \(\mathcal{D}\) takes the fused latent representation
\(\mathbf{F} \in \mathbb{R}^{L_{d} \times D_{c}}\) as input and, through
an upsampling process, reconstructs the motion sequence
\(\widehat{\mathbf{M}} = \mathcal{D}(\mathbf{F}) \in \mathbb{R}^{L \times D}\).

\subsection{Attention-Based Vector Quantization}\label{d.-attention-based-vector-quantization}

Unlike the conventional Euclidean distance--based quantization used in
VQ-VAE \cite{ref2}, which often suffers from codebook collapse, where only a
small subset of codes is selected regardless of the codebook size, we
adopt the attention-based quantization scheme proposed in CoDA \cite{ref15}
to alleviate this issue.

Given an input motion \textbf{M}, we first compute the difference
between the pose latent sequence \({\widehat{\mathbf{z}}}_{1:L_{d}}\)
and the continuous latent sequence \(\mathbf{h}_{1:L_{d}}\) extracted by
the two encoders, and use this difference as the initial residual
\(\mathbf{r}^{(1)} = \left\{ \mathbf{h}_{i} - \mathrm{sg}\left( {\widehat{\mathbf{z}}}_{i} \right) \right\}_{i = 1}^{L_{d}}\),
where \(\mathrm{sg}( \cdot )\) denotes the stop-gradient operator that blocks
gradient backpropagation through its argument. This residual is
progressively quantized over multiple RVQ stages.

At the \emph{s}-th stage, the residual
\(\mathbf{r}^{(s)} \in \mathbb{R}^{L_{d} \times D_{c}}\) is fed into the
\emph{s}-th quantization module \(\mathcal{Q}^{(s)}( \cdot )\), which
outputs a quantized residual
\({\widehat{\mathbf{r}}}^{(s)} = {\widehat{\mathbf{r}}}_{1:L_{d}}^{(s)} \in \mathbb{R}^{L_{d} \times D_{c}}\).
The residual for the next stage is updated as

\begin{equation}
\widehat{\mathbf{r}}^{(s)} = \mathcal{Q}^{(s)}\!\left(\mathbf{r}^{(s)}\right), \qquad
\mathbf{r}^{(s+1)} = \mathbf{r}^{(s)} - \widehat{\mathbf{r}}^{(s)}.
\tag{3}\label{eq:3}
\end{equation}

Let the \emph{s}-th codebook be
\(\mathbf{R}^{(s)} = \left\{ \mathbf{r}_{n}^{(s)} \right\}_{n = 1}^{N_{r}} \subset \mathbb{R}^{D_{c}}\).
From the residual \(\mathbf{r}^{(s)}\) and the codebook entries, we
construct queries and keys using RMSNorm (root-mean-square
normalization) \cite{ref16}:

\begin{equation}
\begin{aligned}
\mathbf{Q} &= \mathrm{RMSNorm}\!\left( \mathbf{r}^{(s)}\mathbf{W}_{Q} \right), \\
\mathbf{K} &= \mathrm{RMSNorm}\!\left( \mathbf{R}^{(s)}\mathbf{W}_{K} \right).
\end{aligned}
\tag{4}\label{eq:4}
\end{equation}

\noindent where
\(\mathbf{W}_{Q} \in \mathbb{R}^{D_{c} \times D_{k}}\),
\(\mathbf{W}_{K} \in \mathbb{R}^{D_{c} \times D_{k}}\), and
\(\mathbf{W}_{V} \in \mathbb{R}^{D_{c} \times D_{v}}\) are learnable
projection matrices for queries, keys, and values, respectively.

Using the scaled dot-product attention between queries and keys, we
obtain both a soft code-selection distribution and a hard one-hot
selection:

\begin{equation}
\begin{aligned}
\mathbf{A}_{\mathrm{soft}} &= \mathrm{softmax}\!\left( \frac{\mathbf{Q}\mathbf{K}^{\top}}{\sqrt{D_{k}}} \right), \\
\mathbf{A}_{\mathrm{hard}} &= \mathrm{hardmax}\!\left( \frac{\mathbf{Q}\mathbf{K}^{\top}}{\sqrt{D_{k}}} \right).
\end{aligned}
\tag{5}\label{eq:5}
\end{equation}

\noindent where \(\mathrm{hardmax( \cdot )}\) selects the maximal entry in each row and
converts it into a one-hot vector.

Letting \(\mathbf{V} = {\mathbf{R}^{(s)}\mathbf{W}}_{V}\) be the value
matrix, we obtain the hard-assignment quantized residual and its soft
counterpart by

\begin{equation}
{\widehat{\mathbf{r}}}^{(s)}\  = \ \mathbf{A}_{hard}\mathbf{V},\ \ {\widehat{\mathbf{r}}}_{soft}^{(s)} = \ \mathbf{A}_{soft}\mathbf{V}\mathbf{.}
\tag{6}\label{eq:6}
\end{equation}

The former is used as the quantized residual passed to the next RVQ
stage and to the decoder, whereas the latter is used in the quantization
loss to stabilize training.

Finally, the final latent representation is obtained by combining the
pose latents with the quantized residuals from all \(S\) RVQ stages:

\begin{equation}
\mathbf{F} = \left\{ {\widehat{\mathbf{z}}}_{i} + \sum_{s = 1}^{S}{\widehat{\mathbf{r}}}_{i}^{(s)} \right\}_{i = 1}^{L_{d}} \in \mathbb{R}^{L_{d} \times D_{c}}\
\tag{7}\label{eq:7}
\end{equation}

This representation \textbf{F} is then fed into the decoder
\(\mathcal{D}\) to reconstruct the motion sequence
\(\widehat{\mathbf{M}}\mathcal{= D(}\mathbf{F}) \in \mathbb{R}^{L \times D}\).

\begin{figure*}[!t]
\centering
\includegraphics[width=\textwidth]{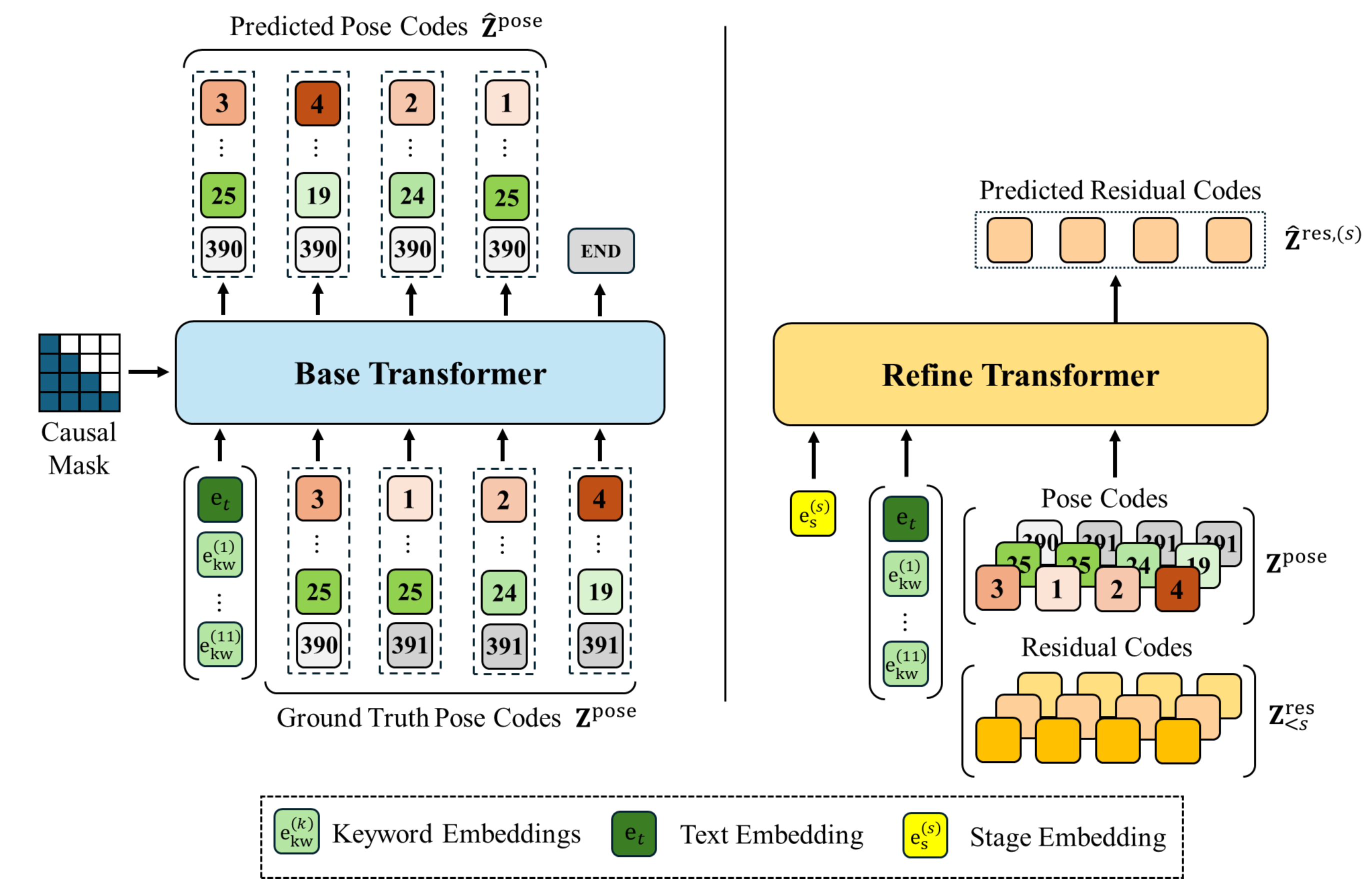}
\caption{Training procedures of the Base Transformer and the Refine Transformer. The Base Transformer autoregressively predicts pose codes conditioned on text, while the Refine Transformer predicts residual codes stage-by-stage conditioned on pose codes, text, and previously predicted residual codes.}
\label{fig:transformers}
\end{figure*}

\subsection{Residual Dropout}\label{e.-residual-dropout}

The pose latent representation is directly aligned with interpretable
categories such as joint angles and relative positions, which makes it
well-suited for manual editing. However, when we simply add the residual
representation to the pose latents, we observe that the model tends to
rely excessively on the residuals to improve reconstruction accuracy,
thereby weakening the controllability provided by the pose codes. To
alleviate this issue, we introduce residual dropout.

At each training step \(t\), we draw a scalar value \(p(t)\) from the
uniform distribution \(\mathcal{U}\lbrack 0,1\rbrack\). If
\(p(t) \geq \tau\), we construct the latent representation \textbf{F} by
adding all quantized residuals
\(\left\{ \sum_{s = 1}^{S}{\widehat{\mathbf{r}}}_{i}^{(s)} \right\}_{i = 1}^{L_{d}}\)to
the pose latent sequence \({\widehat{\mathbf{z}}}_{1:L_{d}}\).
Otherwise, \textbf{F} is formed using only the pose latents
\({\widehat{\mathbf{z}}}_{1:L_{d}}\) without adding any residuals. Here,
\(\tau\) is a threshold that controls the strength of residual dropout.

The reconstructed motion
\(\widehat{\mathbf{M}} = \mathcal{D}(\mathbf{F})\) obtained by feeding
\textbf{F} into the decoder \(\mathcal{D}\) is thus computed while
residuals are stochastically masked during training. This prevents the
model from depending too heavily on the residual representation and
encourages it to preserve both the semantic alignment and the
editability of the pose codes.

\subsection{Base Transformer}\label{f.-base-transformer}

Following CoMo \cite{ref11}, we adopt a decoder-only Base Transformer that
autoregressively predicts pose codes conditioned on text, as illustrated
in the left part of Fig. 4. As text input, we use the motion description
together with 10 body-part keywords and one emotion keyword generated by
GPT-4 \cite{ref17}. These texts are fed into the CLIP text encoder \cite{ref18}
to obtain a sentence embedding \(\mathbf{e}_{t}\) and keyword embeddings
\(\left\{ e_{kw}^{(k)} \right\}_{k = 1}^{11}\), which serve as
conditioning signals.

The target sequence is the K-hot pose code sequence
\(\mathbf{Z}^{pose}\) obtained by passing the given motion through the
pose parser:

\begin{equation}
\mathbf{Z}^{pose} = {\{\mathbf{z}_{i}\}}_{i = 1}^{L_{d}},\ \ \mathbf{z}_{i} = \left\{ z_{i}^{n} \right\}_{n = 1}^{N} \in \left\{ 0,\ 1 \right\}^{N}.\
\tag{8}\label{eq:8}
\end{equation}

\noindent Here, \(z_{i}^{n} \in \left\{ 0,1 \right\}\) is an indicator that
specifies whether the \emph{n}-th pose code \(\mathbf{c}_{n}\) in the
pose codebook is activated at time step \(i\). An additional END token
is appended to indicate the end of the motion, so the actual
dimensionality becomes \(N + 1\).

The Base Transformer employs causal attention and, at each time step
\(i\), takes as input the past pose codes
\(\mathbf{Z}_{1:i - 1}^{pose} = \{\mathbf{z}_{1},...,\mathbf{z}_{i - 1}\}\)
together with the text condition. It models each activation
\(z_{i}^{n}\) as an independent Bernoulli variable and is trained as an
autoregressive multi-label prediction model:

\begin{equation}
P(\mathbf{Z}^{pose}|t) = \prod_{i = 1}^{L_{d}\ }{\prod_{n = 1}^{N + 1}{p_{\theta}\left( z_{i}^{n} \middle| t,z_{1:i - 1}^{1:N + 1} \right)}.}
\tag{9}\label{eq:9}
\end{equation}

\subsection{Refine Transformer}\label{g.-refine-transformer}

Prior work \cite{ref19} has shown that prediction quality can be improved by
feeding the output of a previous stage into the next stage. We extend
this idea and design a Refine Transformer that models the conditional
distribution of residual codes, as depicted on the right side of Fig. 4.

The Refine Transformer is an encoder-only Transformer that takes as
input the text condition \(t\), the pose-code sequence
\(\mathbf{Z}^{pose}\), the target quantization stage index \(s\), and
the residual codes from all preceding stages \(\mathbf{Z}_{< s}^{res}\).
It autoregressively predicts the residual code sequence at stage \(s\),

\begin{equation}
\mathbf{Z}^{res,(s)} = \left\{ \mathbf{z}_{i}^{res,\ (s)} \right\}_{i = 1}^{L_{d}} \subset \mathbb{R}^{N_{r}}.
\tag{10}\label{eq:10}
\end{equation}

Using the self-attention mechanism \cite{ref6}, the model jointly considers
the interactions among pose codes, the text condition, the target stage
index, and the previously predicted residual codes. Over all RVQ stages
\(s = 1,\ldots,S\), it is trained to model the conditional distribution
of residual codes as

\begin{equation}
\begin{split}
P\!\left( \mathbf{Z}^{\mathrm{res},(1):(S)} \,\middle|\, \mathbf{Z}^{\mathrm{pose}}, \mathbf{Z}^{\mathrm{res}}_{<s}, s, t \right) \\
= \prod_{s=1}^{S}\ \prod_{i=1}^{L_d}
p_{\theta}\!\left( \mathbf{z}^{\mathrm{res},(s)}_{i} \,\middle|\, \mathbf{Z}^{\mathrm{pose}}, \mathbf{Z}^{\mathrm{res}}_{<s}, s, t \right).
\end{split}
\tag{11}\label{eq:11}
\end{equation}

\subsection{Inference}\label{h.-inference}

During inference, the Base Transformer first generates a pose-code
sequence autoregressively from the text condition \(t\) until an END
token is produced. The predicted pose codes, together with the text
condition, stage index, and residual codes from previous stages, are
then fed into the Refine Transformer, which sequentially predicts the
residual code sequence for each of the \(S\) quantization stages.
Finally, the pose codes and residual codes are summed frame-wise to form
the latent representation, which is passed to the decoder to synthesize
the final motion.

\section{Training Objective Details}\label{sec:training}

\subsection{Feature Matching Loss}\label{a.-feature-matching-loss}

The pose-guided RVQ tokenizer learns latent representations by
reconstructing the input motion \textbf{M} as a reconstructed motion
\(\widehat{\mathbf{M}}\). Equation (12) defines the motion
reconstruction loss, which measures the difference between \textbf{M}
and \(\widehat{\mathbf{M}}\) in joint space.

\begin{equation}
\mathcal{L}_{motion} = \left\| \mathbf{M} - \widehat{\mathbf{M}} \right\|_{1}
\tag{12}\label{eq:12}
\end{equation}

\subsection{Quantization Loss}\label{b.-quantization-loss}

The quantization loss \(\mathcal{L}_{rvq}\) in (14) follows the
formulation in \cite{ref20}. For each stage \(s\), the per-stage loss
\(\mathcal{L}_{rvq}^{(s)}\) consists of three terms: a reconstruction
term between the input residual \(\mathbf{r}^{(s)}\) and the quantized
residual \({\widehat{\mathbf{r}}}^{(s)}\), a commitment term that
updates the residual codebook, and a term that encourages the soft
assignment \({\widehat{\mathbf{r}}}_{soft}^{(s)}\) to be close to the input
residual:

\begin{equation}
\begin{split}
\mathcal{L}_{rvq}^{(s)}
&= \left\| \mathrm{sg}\!\left(\widehat{\mathbf{r}}^{(s)}\right) - \mathbf{r}^{(s)} \right\|_{2}^{2} \\
&\quad + \beta \left\| \widehat{\mathbf{r}}^{(s)} - \mathrm{sg}\!\left(\mathbf{r}^{(s)}\right) \right\|_{2}^{2}
      + \left\| \widehat{\mathbf{r}}_{soft}^{(s)} - \mathbf{r}^{(s)} \right\|_{2}^{2}.
\end{split}
\tag{13}\label{eq:13}
\end{equation}

\noindent where \(\text{sg}( \cdot )\) denotes the stop-gradient operator and
\(\beta\) is a weighting factor for the commitment term. The overall
quantization loss is then given by the average over all \(S\) stages:

\begin{equation}
\mathcal{L}_{rvq} = \frac{1}{S}\sum_{s = 1}^{S}\mathcal{L}_{rvq}^{(s)}.
\tag{14}\label{eq:14}
\end{equation}

\subsection{Entropy Loss}\label{c.-entropy-loss}

To prevent codebook collapse and promote diverse code usage during
quantization, we adopt the entropy-based loss \(\mathcal{L}_{ent}\) in
(15), following \cite{ref20}:

\begin{equation}
\mathcal{L}_{ent} = \gamma\frac{1}{S}\sum_{s = 1}^{S}\Big(
\mathbb{E}\!\left[ H\!\left( p_{r}^{(s)} \right) \right]
- H\!\left[ \mathbb{E}\!\left( p_{r}^{(s)} \right) \right]
\Big).
\tag{15}\label{eq:15}
\end{equation}

\noindent Here,
\(\mathbb{E}\left\lbrack H\left( p_{r}^{(s)} \right) \right\rbrack\)
denotes the average entropy of the sample-wise code selection
distributions \(p_{r}^{(s)}\) at stage \(s\); minimizing this term
encourages confident (low-entropy) selections for each sample. In
contrast,
\(H\left\lbrack \mathbb{E}\left( p_{r}^{(s)} \right) \right\rbrack\) is
the entropy of the batch-averaged code selection distribution
\(\mathbb{E\lbrack}p_{r}^{(s)}\rbrack\); maximizing this entropy (via
the negative sign) encourages a balanced use of different codes rather
than collapsing to a small subset. The loss is averaged over stages, and
\(\gamma\) is a weighting coefficient.

\subsection{Final Loss}\label{d.-final-loss}

The pose-guided RVQ tokenizer is trained with the combined reconstruction loss in (16),
which consists of the motion reconstruction loss, the RVQ quantization loss, and the entropy loss:

\begin{equation}
\mathcal{L}_{rec} = \mathcal{L}_{motion} + \mathcal{L}_{rvq} + \mathcal{L}_{ent}
\tag{16}\label{eq:16}
\end{equation}

The Base Transformer is trained using the binary cross-entropy loss in
(17), which minimizes the average negative log-likelihood of the pose
codes:

% =====================
% Tables placement request:
% - Page 7: Tables I--III
% - Page 8: Table IV (KIT-ML)
% =====================

\begin{table*}[!t]
\caption{Quantitative results on the HumanML3D~\cite{ref1} test set. Ours (base) denotes using only the Base Transformer, while Ours denotes using both the Base and Refine Transformers. The best performance is highlighted in bold and the second-best is underlined.}
\label{tab:humanml3d}
\centering
\scriptsize
\renewcommand{\arraystretch}{1.25}
\setlength{\extrarowheight}{1pt}

\resizebox{\textwidth}{!}{%
\begin{tabular}{@{\hspace{2pt}} l|ccc|c|c|c|c|c @{\hspace{2pt}}}
\hline
\multirow{2}{*}{Methods} & \multicolumn{3}{c|}{R-Precision $\uparrow$} & \multirow{2}{*}{FID $\downarrow$} & \multirow{2}{*}{MM-Dist $\downarrow$} & \multirow{2}{*}{Diversity $\uparrow$} & \multirow{2}{*}{\shortstack{Multi-\\Modality$\uparrow$}} & \multirow{2}{*}{Editable} \\
\cline{2-4}
 & Top 1 & Top 2 & Top 3 &  &  &  &  &  \\
\hline
Real Motion    & \pmfmt{0.511}{.003} & \pmfmt{0.703}{.003} & \pmfmt{0.797}{.002} & \pmfmt{0.002}{.000} & \pmfmt{2.974}{.008} & \pmfmt{9.503}{.085} & -- & -- \\
CoMo Recons.   & \pmfmt{0.508}{.002} & \pmfmt{0.697}{.002} & \pmfmt{0.792}{.009} & \pmfmt{0.041}{.000} & \pmfmt{3.003}{.006} & \bpm{9.563}{.100} & -- & Yes \\
Ours Recons.   & \bpm{0.514}{.003} & \bpm{0.702}{.002} & \bpm{0.796}{.002} & \bpm{0.007}{.000} & \bpm{2.968}{.005} & \pmfmt{9.555}{.072} & -- & Yes \\
\hline
Guo et al.~\cite{ref1}     & \pmfmt{0.457}{.002} & \pmfmt{0.639}{.003} & \pmfmt{0.740}{.003} & \pmfmt{1.067}{.002} & \pmfmt{3.340}{.008} & \pmfmt{9.188}{.002} & \pmfmt{2.090}{.083} & No \\
TM2T~\cite{ref3}           & \pmfmt{0.424}{.002} & \pmfmt{0.618}{.002} & \pmfmt{0.729}{.002} & \pmfmt{1.501}{.017} & \pmfmt{3.467}{.011} & \pmfmt{8.589}{.086} & \ulpm{2.424}{.093} & No \\
MLD~\cite{ref8}            & \pmfmt{0.481}{.003} & \pmfmt{0.673}{.003} & \pmfmt{0.772}{.002} & \pmfmt{0.473}{.013} & \pmfmt{3.196}{.010} & \ulpm{9.724}{.082} & \pmfmt{1.533}{.008} & No \\
T2M-GPT~\cite{ref4}        & \pmfmt{0.491}{.001} & \pmfmt{0.680}{.003} & \pmfmt{0.775}{.002} & \bpm{0.116}{.004} & \pmfmt{3.118}{.011} & \bpm{9.761}{.081} & \pmfmt{1.831}{.048} & No \\
MotionGPT~\cite{ref5}      & \pmfmt{0.492}{.003} & \pmfmt{0.681}{.003} & \pmfmt{0.778}{.002} & \pmfmt{0.232}{.008} & \pmfmt{3.096}{.008} & \pmfmt{9.528}{.071} & \pmfmt{2.008}{.084} & No \\
GraphMotion~\cite{ref9}    & \ulpm{0.504}{.003} & \ulpm{0.699}{.009} & \ulpm{0.785}{.002} & \ulpm{0.116}{.007} & \ulpm{3.070}{.008} & \pmfmt{9.692}{.067} & \bpm{2.766}{.096} & No \\
\hline
MDM~\cite{ref7}            & \pmfmt{0.320}{.005} & \pmfmt{0.498}{.004} & \pmfmt{0.611}{.007} & \pmfmt{0.544}{.044} & \pmfmt{5.566}{.027} & \pmfmt{9.559}{.086} & \ulpm{2.799}{.072} & Yes \\
FineMoGen~\cite{ref10}     & \ulpm{0.504}{.003} & \ulpm{0.690}{.002} & \ulpm{0.784}{.002} & \pmfmt{0.151}{.008} & \bpm{2.998}{.008} & \pmfmt{9.263}{.067} & \ulpm{2.696}{.079} & Yes \\
CoMo~\cite{ref11}          & \ulpm{0.502}{.002} & \bpm{0.692}{.007} & \bpm{0.790}{.002} & \pmfmt{0.262}{.004} & \ulpm{3.032}{.015} & \bpm{9.936}{.066} & \pmfmt{1.013}{.046} & Yes \\
Ours (Base)                & \pmfmt{0.489}{.002} & \pmfmt{0.681}{.003} & \pmfmt{0.779}{.003} & \pmfmt{0.308}{.008} & \pmfmt{3.070}{.007} & \ulpm{9.904}{.081} & \pmfmt{0.980}{.019} & Yes \\
Ours                       & \pmfmt{0.488}{.002} & \pmfmt{0.677}{.002} & \pmfmt{0.775}{.002} & \ulpm{0.172}{.007} & \pmfmt{3.070}{.007} & \pmfmt{9.588}{.093} & \pmfmt{1.170}{.022} & Yes \\
\hline
\end{tabular}%
}
\end{table*}

\begin{table*}[!t]
\caption{Effect of quantization strategy and codebook utilization on motion reconstruction performance.}
\label{tab:quant_strategy}
\centering
\renewcommand{\arraystretch}{1.25}
\setlength{\extrarowheight}{1pt}
\setlength{\tabcolsep}{6pt}
\begin{tabular}{l c c c c c}
\hline
Methods & Quantization & Perplexity $\uparrow$ & Top 1 R-precision $\uparrow$ & FID $\downarrow$ & MM-Dist $\downarrow$ \\
\hline
CoMo~\cite{ref11} & -- & -- & \pmfmt{0.508}{.002} & \pmfmt{0.041}{.000} & \pmfmt{3.003}{.006} \\
\hline
\multirow{2}{*}{Ours} & Distance-based  & 278.73 & \pmfmt{0.511}{.001} & \pmfmt{0.009}{.000} & \pmfmt{2.982}{.006} \\
                      & Attention-based & \bval{314.12} & \bpm{0.514}{.003} & \bpm{0.007}{.000} & \bpm{2.968}{.005} \\
\hline
\end{tabular}
\end{table*}

\begin{table*}[!t]
\caption{Ablation study on residual dropout. ``w/o residual dropout'' denotes model trained without residual dropout.}
\label{tab:ablation_dropout}
\centering
\renewcommand{\arraystretch}{1.25}
\setlength{\extrarowheight}{1pt}
\setlength{\tabcolsep}{6pt}
\begin{tabular}{l c c c c}
\hline
Methods & Orthogonality $\downarrow$ & Top 1 R-precision $\uparrow$ & FID $\downarrow$ & MM-Dist $\downarrow$ \\
\hline
CoMo~\cite{ref11} & \bval{0.005} & \pmfmt{0.508}{.002} & \pmfmt{0.041}{.000} & \pmfmt{3.003}{.006} \\
Ours (w/o residual dropout) & 0.045 & \pmfmt{0.510}{.003} & \bpm{0.007}{.000} & \pmfmt{2.981}{.008} \\
Ours ($\tau = 0.1$) & 0.022 & \bpm{0.514}{.003} & \bpm{0.007}{.000} & \bpm{2.968}{.005} \\
Ours ($\tau = 0.5$) & 0.011 & \pmfmt{0.510}{.002} & \pmfmt{0.014}{.000} & \pmfmt{2.980}{.006} \\
\hline
\end{tabular}
\end{table*}

\begingroup
\setlength{\abovedisplayskip}{4pt}
\setlength{\belowdisplayskip}{4pt}
\setlength{\abovedisplayshortskip}{4pt}
\setlength{\belowdisplayshortskip}{4pt}

\begin{equation}
\begin{aligned}
\mathcal{L}_{base} &= \\
&&\hspace{-50pt}-\frac{1}{L(N+1)} \sum_{i=1}^{L}\sum_{n=1}^{N+1}
\mathbb{E}_{z_i^n \sim \mathrm{Ber}(z_i^n)}
\left[ \log p_{\theta}\!\left( z_i^n \mid t, z_{1:i-1}^{1:N+1} \right) \right].
\end{aligned}
\tag{17}\label{eq:17}
\end{equation}

\endgroup

The Refine Transformer is trained with the cross-entropy loss in (18),
which minimizes the average negative log-likelihood of the residual
codes conditioned on the pose codes, text, stage index, and previous
residual codes:

\begingroup
\setlength{\abovedisplayskip}{4pt}
\setlength{\belowdisplayskip}{4pt}
\setlength{\abovedisplayshortskip}{4pt}
\setlength{\belowdisplayshortskip}{4pt}
\begin{equation}
\mathcal{L}_{ref} =
-\frac{1}{S L_{d}}\sum_{s=1}^{S}\sum_{i=1}^{L_{d}}
\left[ \log p_{\phi}\!\left( \mathbf{z}_{i}^{res,(s)} \mid \mathbf{Z}^{pose}, \mathbf{Z}_{<s}^{res}, s, t \right) \right].
\tag{18}\label{eq:18}
\end{equation}
\endgroup

% (Table IV should float to the top of the next page)

\begin{table*}[!t]
\caption{Quantitative results on the KIT-ML~\cite{ref21} test set. Ours (base) denotes using only the Base Transformer, while Ours denotes using both the Base and Refine Transformers. The best performance is highlighted in bold and the second-best is underlined. $\dagger$ indicates results obtained without using fine-grained keywords.}
\label{tab:kitml}
\centering
\scriptsize
\renewcommand{\arraystretch}{1.25}
\setlength{\extrarowheight}{1pt}

\resizebox{\textwidth}{!}{%
\begin{tabular}{@{\hspace{2pt}} l|ccc|c|c|c|c|c @{\hspace{2pt}}}
\hline
\multirow{2}{*}{Methods} & \multicolumn{3}{c|}{R-Precision $\uparrow$} & \multirow{2}{*}{FID $\downarrow$} & \multirow{2}{*}{MM-Dist $\downarrow$} & \multirow{2}{*}{Diversity $\uparrow$} & \multirow{2}{*}{\shortstack{Multi-\\Modality$\uparrow$}} & \multirow{2}{*}{Editable} \\
\cline{2-4}
 & Top 1 & Top 2 & Top 3 &  &  &  &  &  \\
\hline
Real Motion    & \pmfmt{0.424}{.003} & \pmfmt{0.649}{.003} & \pmfmt{0.779}{.002} & \pmfmt{0.031}{.000} & \pmfmt{2.788}{.008} & \pmfmt{11.08}{.097} & -- & -- \\
CoMo Recons.   & \pmfmt{0.387}{.005} & \pmfmt{0.603}{.005} & \pmfmt{0.730}{.005} & \pmfmt{0.254}{.007} & \pmfmt{3.046}{.011} & \pmfmt{10.73}{.128} & -- & Yes \\
Ours Recons.   & \bpm{0.415}{.004} & \bpm{0.639}{.006} & \bpm{0.771}{.006} & \bpm{0.083}{.002} & \bpm{2.824}{.013} & \bpm{10.97}{.002} & -- & Yes \\
\hline
Guo et al.~\cite{ref1}     & \pmfmt{0.370}{.005} & \pmfmt{0.569}{.007} & \pmfmt{0.693}{.007} & \pmfmt{2.770}{.109} & \pmfmt{3.401}{.008} & \pmfmt{10.91}{.119} & \pmfmt{1.482}{.065} & No \\
TM2T~\cite{ref3}           & \pmfmt{0.280}{.002} & \pmfmt{0.463}{.006} & \pmfmt{0.587}{.005} & \pmfmt{3.599}{.153} & \pmfmt{4.591}{.026} & \pmfmt{9.473}{.117} & \ulpm{3.292}{.081} & No \\
MLD~\cite{ref8}            & \pmfmt{0.390}{.008} & \pmfmt{0.609}{.007} & \pmfmt{0.734}{.007} & \ulpm{0.404}{.027} & \pmfmt{3.204}{.027} & \pmfmt{10.80}{.117} & \pmfmt{2.192}{.065} & No \\
T2M-GPT~\cite{ref4}        & \ulpm{0.416}{.006} & \ulpm{0.627}{.006} & \ulpm{0.745}{.006} & \pmfmt{0.514}{.029} & \ulpm{3.007}{.023} & \pmfmt{10.92}{.108} & \pmfmt{1.570}{.039} & No \\
MotionGPT~\cite{ref5}      & \pmfmt{0.366}{.005} & \pmfmt{0.558}{.004} & \pmfmt{0.680}{.005} & \pmfmt{0.510}{.016} & \pmfmt{3.527}{.021} & \pmfmt{10.35}{.084} & \pmfmt{2.328}{.117} & No \\
GraphMotion~\cite{ref9}    & \bpm{0.429}{.007} & \bpm{0.648}{.006} & \bpm{0.769}{.008} & \bpm{0.313}{.013} & \pmfmt{3.076}{.022} & \bpm{11.12}{.135} & \bpm{3.627}{.113} & No \\
\hline
MDM~\cite{ref7}            & \pmfmt{0.164}{.004} & \pmfmt{0.291}{.004} & \pmfmt{0.396}{.004} & \pmfmt{0.497}{.021} & \pmfmt{9.191}{.022} & \pmfmt{10.85}{.109} & \bpm{1.907}{.214} & Yes \\
FineMoGen~\cite{ref10}     & \bpm{0.432}{.006} & \bpm{0.649}{.005} & \pmfmt{0.772}{.008} & \bpm{0.178}{.007} & \bpm{2.869}{.014} & \pmfmt{10.85}{.115} & \ulpm{1.877}{.093} & Yes \\
CoMo~\cite{ref11}          & \pmfmt{0.422}{.009} & \pmfmt{0.638}{.007} & \ulpm{0.765}{.011} & \pmfmt{0.332}{.045} & \ulpm{2.873}{.021} & \pmfmt{10.95}{.196} & \pmfmt{1.249}{.008} & Yes \\
CoMo$^{\dagger}$~\cite{ref11} & $0.399$ & -- & -- & $0.399$ & $2.898$ & \bval{11.26} & -- & Yes \\
Ours$^{\dagger}$ (Base)    & \pmfmt{0.387}{.006} & \pmfmt{0.594}{.006} & \pmfmt{0.725}{.005} & \pmfmt{0.708}{.034} & \pmfmt{3.042}{.022} & \pmfmt{10.89}{.113} & \pmfmt{1.137}{.044} & Yes \\
Ours$^{\dagger}$           & \pmfmt{0.424}{.007} & \ulpm{0.639}{.008} & \pmfmt{0.757}{.006} & \ulpm{0.328}{.018} & \pmfmt{2.875}{.026} & \ulpm{10.99}{.196} & \pmfmt{1.376}{.051} & Yes \\
\hline
\end{tabular}%
}
\end{table*}

\section{Experiments}\label{sec:experiments}

\subsection{Dataset}\label{a.-dataset}

We conduct experiments on two text-to-motion datasets, HumanML3D \cite{ref1}
and KIT-ML \cite{ref21}. HumanML3D \cite{ref1} consists of 14,616 motion
sequences collected from AMASS \cite{ref22} and HumanAct12 \cite{ref23} and
44,970 corresponding textual descriptions. The motions are normalized to
20 fps and augmented via left--right mirroring. KIT-ML \cite{ref21} contains
3,911 motion sequences and 6,279 text descriptions, built from KIT
\cite{ref24} and CMU \cite{ref25} motion capture data and normalized to 12.5
fps. Both datasets are split into training, validation, and test sets
with ratios of 80\%, 5\%, and 15\%, respectively.

\subsection{Experimental Setup}\label{b.-experimental-setup}

For each motion sequence, we perform data augmentation by cropping a
temporal window of 64 frames centered at a randomly selected frame and
then downsampling the sequence by a factor of 4 along the temporal axis.
The RVQ module consists of six stages, each operating on 512-dimensional
vectors. We use a pose codebook with 392 codes and a residual codebook
with 512 codes, both of dimension 512. During training, the batch size
is set to 256 and the learning rate to \(2 \times 10^{- 4}\) with a
linear warm-up over the first 1,000 steps. The loss-weight
hyperparameters are \(\beta = 0.25\) and \(\gamma = 0.01\), and the
residual dropout threshold is fixed to \(\tau = 0.1\). All experiments
are conducted on an NVIDIA A100-SXM4-80GB GPU.

\subsection{Evaluation Metrics}\label{c.-evaluation-metrics}

We follow the evaluation protocol of T2M-GPT \cite{ref4} and report Fr\'echet
inception distance (FID), R-Precision, multimodal distance (MM-DIST),
diversity, and multimodality. FID measures the similarity between real
and generated motion distributions. R-Precision and MM-DIST evaluate how
well a generated motion sequence matches the corresponding text
description. Diversity and multimodality assess, respectively, the
variety of generated motion sequences and the diversity of motions that
can be produced under the same text condition.

All these metrics are computed using motion and text features extracted
from the pretrained model provided in \cite{ref1}. In addition, we introduce
Perplexity and Orthogonality. Perplexity measures the utilization of the
residual codebook and is defined as

\begin{equation}
\mathrm{Perplexity} = \frac{1}{S}\sum_{s = 1}^{S}{\exp\left( - \sum_{i = 1}^{N_{r}}{p_{i}^{(s)}\log\left( p_{i}^{(s)} \right)} \right)}.
\tag{19}\label{eq:19}
\end{equation}

\noindent where \(p_{i}^{(s)}\) denotes the code usage distribution at the \(s\)-th
quantization stage, obtained by averaging the one-hot vectors of the
selected codes.

Orthogonality evaluates how well pose codes are separated from each
other based on their inner-product similarity, and is computed as

\begin{equation}
\mathrm{Orthogonality}
= \frac{1}{N(N-1)}
\sum_{\substack{i,j=1\\ i\neq j}}^{N}
\left\langle
\frac{\mathbf{c}_i}{\lVert \mathbf{c}_i\rVert},
\frac{\mathbf{c}_j}{\lVert \mathbf{c}_j\rVert}
\right\rangle^{2}
\tag{20}\label{eq:20}
\end{equation}

\subsection{Generation Results}\label{d.-generation-results}

For model selection, we use the checkpoint that achieves the lowest FID
on the validation set. Tables I and IV report the mean and 95\%
confidence intervals over 20 evaluation runs. On both HumanML3D \cite{ref1}
and KIT-ML \cite{ref21}, the proposed method reduces the FID for both
reconstruction and generation compared with CoMo \cite{ref11}. MDM \cite{ref7} and
FineMoGen \cite{ref10} must resample motions by re-running the diffusion-based
generative pipeline for editing operation, which makes it difficult to
guarantee consistency between the pre- and post-edit motions. In
contrast, CoMo \cite{ref11} proposed a pose code--based,
representation-level editing framework that preserves the structural
consistency of the original motion. Building on this idea, our method
maintains the same representation-level editing structure while further
reducing FID compared with CoMo \cite{ref11}, thereby improving motion quality. On
KIT-ML \cite{ref21}, it also improves Top-R-Precision even without fine-grained
keywords, indicating better overall motion quality and text--motion
alignment. Consequently, our approach establishes a strong baseline
within interpretable, representation-level motion editing by improving
FID over CoMo \cite{ref11}, and is conceptually distinct from
diffusion-based generative models such as MDM \cite{ref7} and FineMoGen \cite{ref10}.

Table II shows that attention-based quantization outperforms Euclidean
distance--based quantization: it achieves higher codebook utilization
for motion reconstruction and better FID, Top-R, and MM-DIST scores. Table III reports the effect of residual dropout and the threshold
\(\tau\); \(\tau = 0.1\) provides the best trade-off between FID and
Orthogonality, yielding the most balanced reconstruction performance.

% [moved tab:rd table]

These quantitative improvements are consistent with the qualitative results shown in Fig.~5, the first row shows motions generated from the text
prompt \emph{``the soccer player kicks the ball.''} While CoMo \cite{ref11}
produces a relatively limited range of leg movements, our model
generates motions with a wider range of motion. In the second row, for
the prompt \emph{``a person runs forward then abruptly turns to the left
and continues running,''} the proposed model more effectively captures
details such as the sudden change in velocity compared with CoMo
\cite{ref11}. These results demonstrate that temporal variations and subtle
motion details that cannot be represented by pose codes alone are
effectively captured through the residual codes.

\begin{figure}[!t]
\centering
\includegraphics[width=\columnwidth]{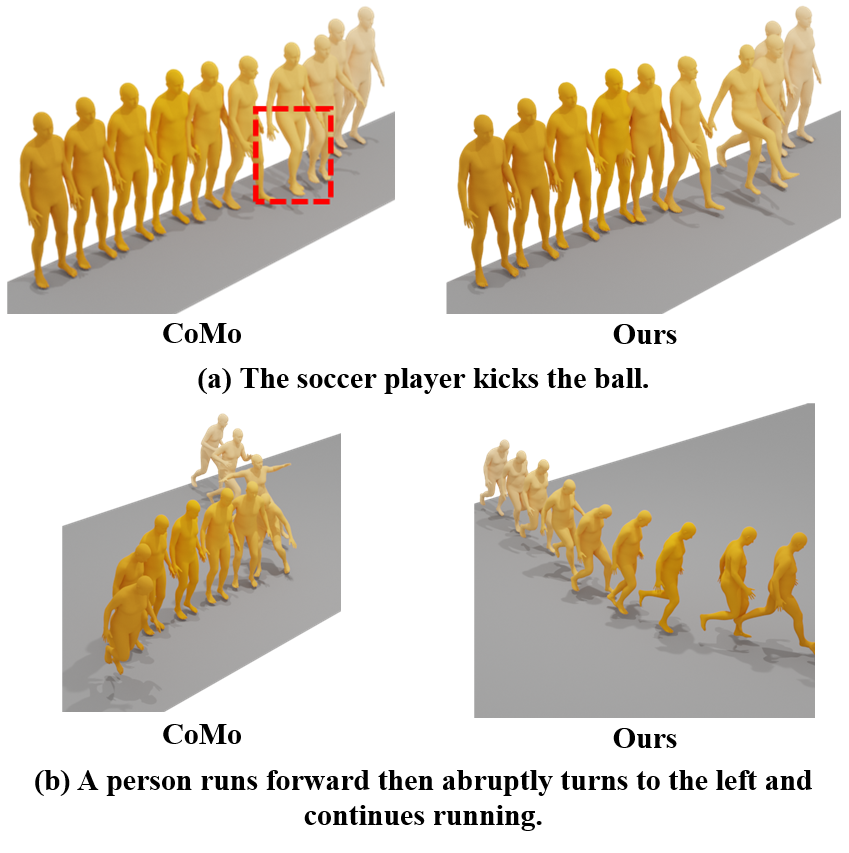}
\caption{Qualitative comparison of motion generation. (a) Prompt: ``the soccer player kicks the ball.'' (b) Prompt: ``a person runs forward then abruptly turns to the left and continues running.''}
\label{fig:qual_gen}
\end{figure}

\begin{figure}[!t]
\centering
\includegraphics[width=\columnwidth]{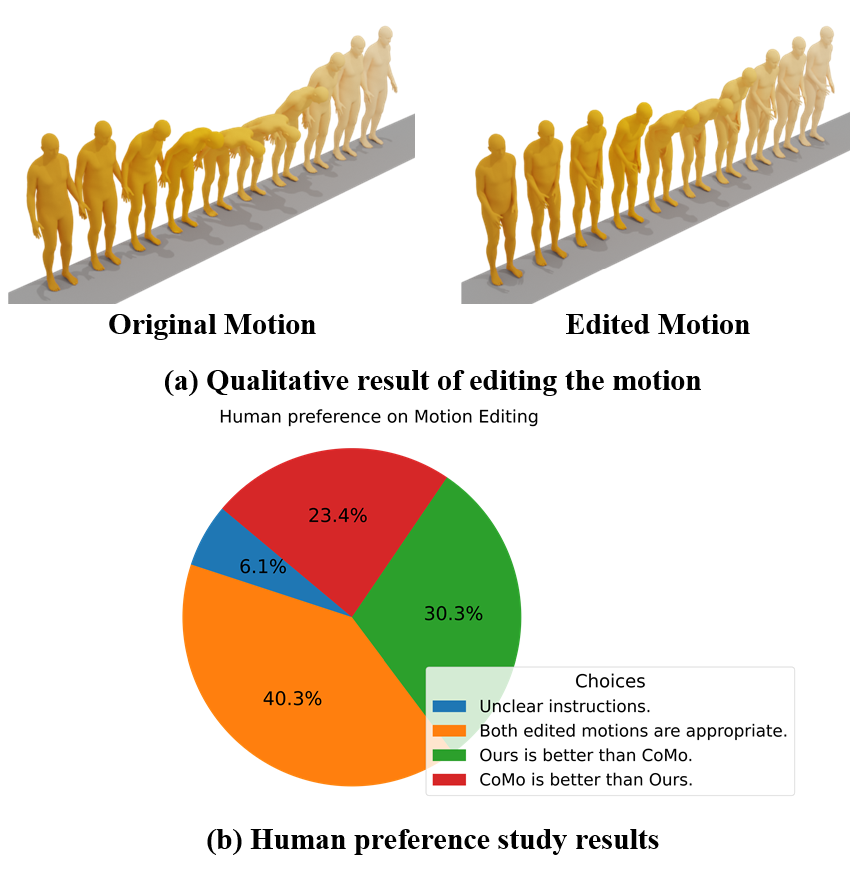}
\caption{(a) Qualitative result of editing a motion according to text instruction. (b) Human preference study results comparing methods in terms of motion quality and edit consistency.}
\label{fig:user_study}
\end{figure}

\subsection{Editing Results}\label{e.-editing-results}

We follow the editing pipeline of CoMo \cite{ref11} by modifying pose codes to
perform motion editing and then conduct a user preference study on the results.
The top part of Fig.~6 illustrates an example where GPT-4 \cite{ref17} suggests
pose codes and a temporal segment to be edited for the motion
\emph{``this person bends forward as if to bow.''} We then modify the pose so that
\emph{``the hands are placed slightly closer to the body.''} The bottom part of
Fig.~6 reports the preferences of 21 undergraduate participants over 11 edited
motion samples, indicating that the proposed model enables meaningful motion
editing while preserving edit consistency.

\section{Conclusion}\label{sec:conclusion}

In summary, pose-guided residual refinement for motion (PGR²M) enhances
pose-code-based motion representations by structurally combining
interpretable pose codes with temporally sensitive residual codes. By
introducing RVQ-based residual modeling and residual dropout, the model
preserves the semantics of pose codes while effectively capturing subtle
temporal variations and high-frequency motion details that are difficult
to express with pose codes alone. Furthermore, the Refine Transformer
predicts residual codes conditioned on text, pose codes, and stage
indices, complementing the information missing from pose codes.
Together, these components make PGR²M a flexible and extensible base
model for human motion synthesis that simultaneously offers
interpretability, controllability, and high fidelity.

\end{document}